\documentclass[1p]{elsarticle}

\usepackage[T1]{fontenc} 
\usepackage[utf8]{inputenc}

\usepackage{hyperref}

 \usepackage{numcompress}

\usepackage{graphicx}
\usepackage{comment}
\usepackage{amsfonts}
\usepackage{amsmath}
\usepackage{tabularx}
\usepackage{multirow}
\usepackage{colortbl}
\usepackage[toc,page]{appendix}

\usepackage{tikz-qtree}
\usepackage[edges]{forest}

\usepackage{hyperref}
\hypersetup{
    colorlinks=true,
    linkcolor=blue,
    filecolor=magenta,      
    urlcolor=cyan,
}

\usepackage{booktabs}
\usepackage{verbatim}
\usepackage{adjustbox}

\usepackage{url}
\usepackage{lscape}

\begin{document}

\begin{frontmatter}

\title{Bladder segmentation based on deep learning approaches: current limitations and lessons}



\author[mymainaddress]{Mark G. Bandyk\corref{mycorrespondingauthor}}
\ead{Mark.Bandyk@jax.ufl.edu}
\author[mysecondaryaddress]{Dheeraj R Gopireddy}
\author[mysecondaryaddress]{Chandana Lall}
\author[mymainaddress]{K.C. Balaji}
\author[ETSaddress]{Jose Dolz\corref{mycorrespondingauthor}}

\cortext[mycorrespondingauthor]{Corresponding authors}
\ead{jose.dolz@etsmtl.ca}


\address[mymainaddress]{Department of Urology, University of Florida, Jacksonville, FL}
\address[mysecondaryaddress]{Department of Radiology, University of Florida, Jacksonville, FL}
\address[ETSaddress]{ETS Montreal, QC, Canada}

\begin{abstract}
Precise determination and assessment of bladder cancer (BC) extent of muscle invasion involvement guides proper risk stratification and personalized therapy selection. \textcolor{black}{ In this context, segmentation of both bladder walls and cancer are of pivotal importance, as it provides invaluable information to stage the primary tumor. Hence, multiregion segmentation on patients presenting with symptoms of bladder tumors using deep learning heralds a new level of staging accuracy and prediction of the biologic behavior of the tumor. Nevertheless, despite the success of these models in other medical problems, progress in multiregion bladder segmentation is still at a nascent stage, with just a handful of works tackling a multiregion scenario. Furthermore, most existing approaches systematically follow prior literature in other clinical problems, without casting a doubt on the validity of these methods on bladder segmentation, which may present different challenges. Inspired by this, we} 
provide an in-depth look at bladder cancer segmentation using deep learning models.
The critical determinants for accurate differentiation of muscle invasive disease, \textcolor{black}{current status of deep learning based bladder segmentation, lessons and limitations of prior work are highlighted.}
\end{abstract}

\begin{keyword}
Bladder segmentation, deep learning, convolutional neural networks, bladder cancer.
\end{keyword}

\end{frontmatter}


\section{Introduction}

Bladder cancer is a major global health problem and is the 10th predominant cancer worldwide according to the World Cancer Research Fund \cite{wcrf2018}. Global risk factors include smoking (up to 6 times increased risk, versus non-smokers), parasitic infection (schistosomiasis) and toxic chemicals, such as aromatic amines (occupation exposure) and arsenic (drinking water) \cite{wcrf2018}.  In the United States, bladder cancer is the fourth most common cancer among men, who are four times more likely to contract the disease than women. The median age at diagnosis is 73 years \cite{cancer_org}. The American Cancer Society estimated that there were 81,000 new cases of bladder cancer and 18,000 deaths from bladder cancer in 2020 \cite{cancer_org}.  Early bladder cancer diagnosis and treatment reduces morbidity and mortality and imaging is a key component in the management (3, 4)

\subsection{\textcolor{black}{Limitations of Transurethral resection surgery}}


\textcolor{black}{The precise determination and assessment of the extent of bladder cancer (BC) muscle invasion guides proper risk stratification and personalized therapy selection.  With the diagnosis of muscle invasive bladder cancer (MIBC), the current standard of care is prompt neoadjuvant systemic therapy followed by radical cystectomy and urinary diversion \cite{chang2017treatment,milowsky2016guideline,kulkarni2010updated,sylvester2006predicting}.  The surgical technique of transurethral resection of the bladder tumor (TURBT) enables both the diagnosis and identification of muscle invasion of BC.  Notably, this piecemeal tumor resection has evolved little since its initial description in 1962 \cite{jones1962treatment}, and has potential complications and limitations. Nevertheless, an optimal TURBT is not always performed.  Recent clinical outcome data indicates a high quality TURBT requires experience, clinical judgement, and precise surgical technique \cite{mostafid2012measuring}.  Inadequate sampling of muscle for BC invasion sometimes necessitates a repeat TURBT \cite{naselli2018role}, increasing patient risk and morbidity. A TURBT has up to 6.7\% complications including bladder perforation and uncontrolled bleeding risk \cite{comploj2014perforation}.  Although guidelines recommended a repeat TURBT for improved bladder cancer staging \cite{chang2017treatment}, the benefit of a second TURBT is inconsistent. While repeat resection detects residual tumor in 26 to 83\% of patients \cite{miladi2003value}, occult extravesical cancer cannot be detected by repeat TURBT \cite{karakiewicz2006nomogram}.)  Non-invasive imaging with bladder segmentation is necessary to correct the shortcomings of surgical staging with TURBT. }

\subsection{\textcolor{black}{Limitations of multi-parametric magnetic resonance imaging}}


\textcolor{black}{Multi-parametric magnetic resonance (mp-MR) imaging has been rapidly adopted to evaluate muscle invasion in bladder cancer patients.  The mp-MR imaging allows for high soft tissue contrast resolution and multiplanar imaging, and enables radiologists to predict the depth of tumor invasion.   \cite{huang2018diagnostic,woo2017diagnostic,caglic2020mri,juri2020staging}.  Current mp-MR imaging require improvements in efficiency, accuracy, and consistency to improve BC staging. It lacks reliable discrimination of muscle invasion to reduce or eliminate pathologic staging with TURBT. \cite{juri2020staging,ueno2019diagnostic,barchetti2019multiparametric,del2020prospective}.  Magnetic resonance image evaluations are a tedious, slice by slice process whose effectiveness depends upon the experience of the radiologist. }

\begin{figure}[h!]
    \centering
    \includegraphics[width=1.0\linewidth]{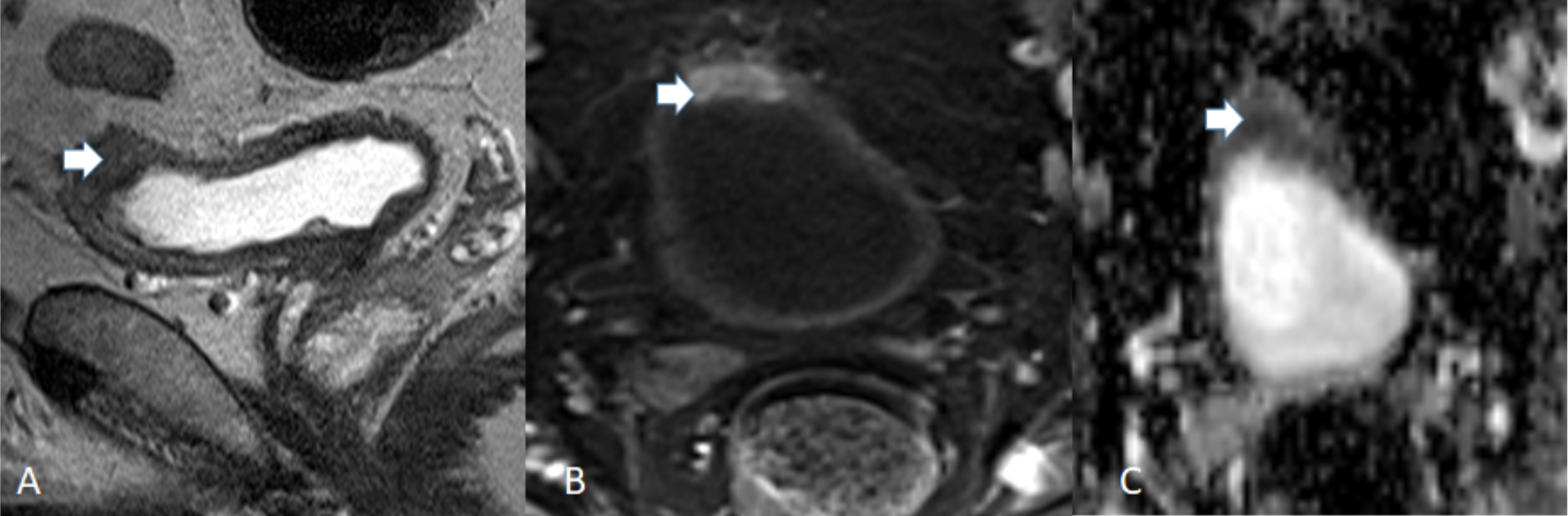}
    \caption{A case of T2 (muscle invasive) bladder cancer. Sagittal T2 weighted MRI image demonstrates a focal area of abnormal signal within the dome of the bladder (arrow) (A). Axial post contrast T1 weighted images show focal full thickness enhancement in the early phase (arrow) (B). Axial ADC diffusion map demonstrates full thickness low ADC signal (arrow) consistent with high grade tumor (C).}
    \label{fig:t2_MIBC}
\end{figure}

\textcolor{black}{Standardized magnetic resonance image (MRI)reporting systems have been adopted for several organ systems including prostate (PI-RADS), breast (BI-RADS) and bladder (VI-RADS). The Vesical Imaging Reporting and Data Systems (VI-RADS) has standardized MRI interpretations and identification of MIBC, the most critical determinate for directing therapy. \cite{panebianco2018multiparametric}.  Using the VI-RADS score for mp-MR images, the first parameter evaluated is T2 weighted-images (T2WI) followed by the diffusion weighted/Apparent Diffusion Coefficient (DWI/ADC) and dynamic-contrast enhanced (DCE) images \cite{panebianco2018multiparametric}. DWI/ADC is the most important image sequence for bladder cancer staging \cite{juri2020staging}. However, DWI/ADC images are prone to motion artifacts, in which case the DCE and T2WI are relied upon to determine MIBC (Figure \ref{fig:t2_MIBC}). In addition, there can be false positive MIBC and over-staging of tumor invasion due to  bladder wall inflammation or fibrosis on T2WI \cite{tekes2005dynamic}. DCE sequence is not helpful in this setting. However, the low sequence intensity on DWI/ADC distinguishes fibrosis from tumor invasion.  In addition to these limitations, VI-RADS lacks validation for patient risk stratification, therapy selection and monitoring therapeutic response. \cite{pecoraro2020overview}.}

\subsection{\textcolor{black}{Accurate bladder imaging guides therapy}}

\begin{figure}[h!]
    \centering
    \includegraphics[width=1.0\linewidth]{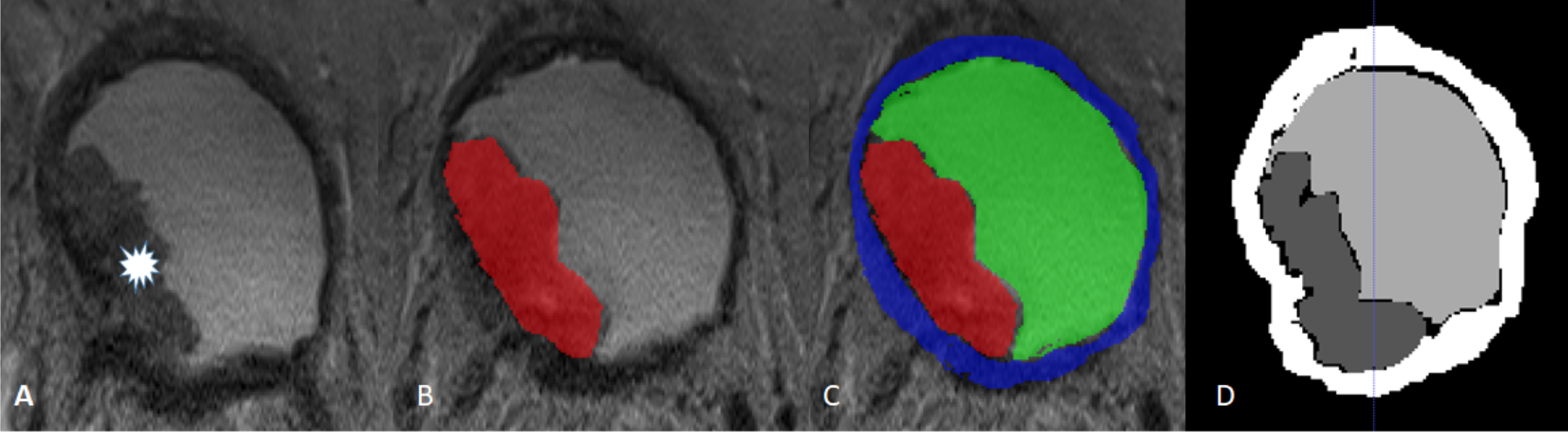}
    \caption{Manual tumor and Bladder wall contouring: Axial T2   weighted MR Images of the bladder (A) with tumor (star) invading the muscle. Systematic contouring of the tumor (B), bladder lumen and wall (C). Segmented mask of the 2 D slice image including tumor, lumen and wall (D).}
    \label{fig:contours}
\end{figure}

\textcolor{black}{Multiparametric magnetic resonance imaging evaluates alterations of the BC mass following therapy, guiding future management.} Criteria from the World Health Organization (WHO) \cite{world1979handbook} and Response Evaluation Criteria in Solid Tumors (RECIST) \cite{eisenhauer2009new} measures the tumor mass response to chemotherapy. WHO criteria defines tumor mass reduction as the percentage reduction using the perpendicular diameters before and after treatment, whereas RECIST criteria uses the percent reduction of the longest diameter to determine therapeutic response.  Unfortunately, both methods are potentially inaccurate due to differences in interpretations among observers, especially in complex and irregular tumors \cite{husband2004evaluation}, like bladder tumors. Advances in the accuracy of three dimensional (3D) deep convolutional neural networks (CNN) algorithms have the potential to automate Gross Tumor Volume (GTV) contouring on multiparametric MRIs, a critical step in staging of bladder cancer. However, there are some innate challenges with regard to the accuracy of tumor contouring (Fig \ref{fig:contours}) which could vary depending on the experience of the radiologist, tumor heterogeneity, poor tumor-to-normal tissue interference and variability in MRI datasets. Substantial successful work has been performed in segmentation of 3D data sets for solid organs like the liver, heart and brain \cite{litjens2017survey}. More recently work by Dolz et al has shown high accuracy with multi region segmentation of bladder cancer through the use of 3D volume T2 sequencing data sets \cite{dolz2018multiregion}. As the viability of deep learning models for the automatic multi region segmentation of bladder cancer MRI images expands, there is an opportunity to utilize this technique to assess response to immunotherapy as well.  

Imaging responses tend to be different following chemotherapy and immunotherapy.  For example, unlike chemotherapy, tumor mass size alterations may be infrequent following immunotherapy. Immunotherapy with checkpoint inhibitors (CPI), a new beneficial class of caner therapy, activates the host immune response.  Unique patterns of the therapeutic immune response seen with CPI and immune-related adverse effects are difficult to measure with the accepted WHO and RECIST criteria \cite{wolchok2009guidelines}.  Traditional chemotherapy targets actively dividing cells, and targeted therapies interfere with cellular molecular events, causing reduction in size of tumor mass.  Immunotherapy response may have a variety of beneficial immune response patterns as well \cite{carter2018immunotherapy}. Atypical therapeutic immune responses will include transient tumor progression, appearance of new lesions, no measurable size alterations or delayed responses \cite{carter2018immunotherapy}. New immune related response criteria have been established, but prospective studies are needed to identify and predict a CPI therapeutic response \cite{weiss2018imaging}. Currently, response to immunotherapy image interpretation depends upon size alteration.  Future image analysis may evaluate therapeutic responses based on tumor heterogeneity and qualitative, not quantitative mass alterations. 

Motivated by this, our main objectives in this work can be summarized as to: 1) shed light about the current status to tackle the task of bladder segmentation based on deep learning, 2) investigate which are the shortcomings of these works compared to recent literature in different medical segmentation problems and 3) give valuable insights on better practices to deploy higher performing models. We anticipate that the lessons provided in this work will become a useful guideline to improve the quality of subsequent similar efforts, dedicated to automating the bladder segmentation task with a deep learning model.

\section{A closer look at convolutional neural networks}
\label{sec:closerlook}

Machine learning, and more recently deep learning, have emerged as powerful tools to assist in the diagnostic, treatment and follow-up of many diseases. Particularly, convolutional neural networks (CNNs) have demonstrated an outstanding performance in visual recognition problems, sometimes surpassing the human level performance. Among these tasks, segmentation of medical images has greatly nourished from the advances on \textcolor{black}{the field of visual recognition based on deep learning}, with remarkable achievements in neuroimaging \cite{dolz20183d,dolz2018hyperdense}, cardiology \cite{bernard2018deep} or oncology \cite{havaei2017brain,sinha2020multi}, among many others. Indeed, these methods dominate the recent literature on these tasks and represent the current state-of-the-art. Let us first introduce the basics before detailing the status of medical image segmentation based on deep learning models.

\subsection{Convolutional neural networks}
Early CNN architectures, like AlexNet \cite{krizhevsky2012imagenet} or VGG-Net \cite{simonyan2014very} were initially designed for image class recognition, which is arguably the simplest and most well known application of CNNs. In this task, an input image is given to the model, which predicts a class of a typically mutually exclusive set of labels for the input image. 
A standard CNN for image classification typically consists of stacks of four types of layers, i.e., convolutional layer, subsampling or pooling layer, fully-connected and a final softmax layer. 
If we take as example a 2-dimensional input image $X$, \textcolor{black}{at each layer this image is convolved with a set of $W$ kernels of size $m \times n$, and a set of biases is added, resulting in a new feature map. Then, a non-linear transformation $f$ is applied to these features maps, which leads to the output of a convolutional layer. This can be expressed as:}

\begin{equation}
   y_{i,j} = f(\sum_{m}\sum_{n}W_{m,n}X_{i+m,j+n}+b_{i,j})
\end{equation}



\textcolor{black}{This process is repeated multiple times, i.e., once per layer, to construct the whole architecture responsible for extracting the image features. Typically, rectified linear units (ReLU) are employed as a transformation function $f$.} 
Afterwards, one or several fully-connected layers are appended to the feature extractor. Unlike convolutional blocks, which share their weights and have sparse connectivity, the neurons in a fully connected layer have a complete connection to all the activations from the previous layers, increasing heavily the number of parameters of the network. The objective of these layers is to learn non-linear combinations of the features extracted by the feature extractor before being fed into the final classification or softmax layer.

The numerical output of the last fully-connected layer is known as 
\textit{logits}. Since these values are not bounded, their direct interpretation is not possible, particularly in a multi-class classification setting. To address this issue, a softmax layer is added at the output of the network, which turns logits into a vector that represents a true probability distribution of potential outputs. The softmax function can be defined as:

\begin{equation}
  s^c_{i,j} = \frac{e^{\hat{y}^c_{i,j}}}{\sum_{c'}^{K} e^{\hat{y}^{c'}_{i,j}}}
\end{equation}

where $\hat{y}^c_{i,j}$ is the logit prediction for the class $c$ at pixel location ${i,j}$, $K$ is the total number of classes, \textcolor{black}{and $s_c$ the softmax probability for the class $c$. Thus, for a problem with $K$ classes, the output at pixel ${i,j}$ is a vector of the form $\mathbf{s}_{i,j}=[s^0_{i,j},...,s^{K-1}_{i,j}]$.}


\textbf{Training a CNN.} Neural networks are trained using the stochastic gradient descent algorithm, where the network weights are updated by backpropagating the output error. In this optimization scenario, the function employed to evaluate a set of weights is typically referred to as the objective function or \textit{loss function}. With softmax as output activation layer, the cross-entropy is the most widely used loss function, which over $N$ training images is defined as:


\begin{equation} \mathcal{L}=-\frac{1}{N}\sum_{t}^N \mathbf{y}_t  \log(\mathbf{s}_t)
\end{equation}


\textcolor{black}{where $\mathbf{s}_t$ is the vector of pixel-wise probability predictions of the $N$ pixels of a given image $t$ and $\mathbf{y}_t$ its corresponding ground truth.}

\begin{figure}[h!]
    \centering
    \includegraphics[width=0.90\linewidth]{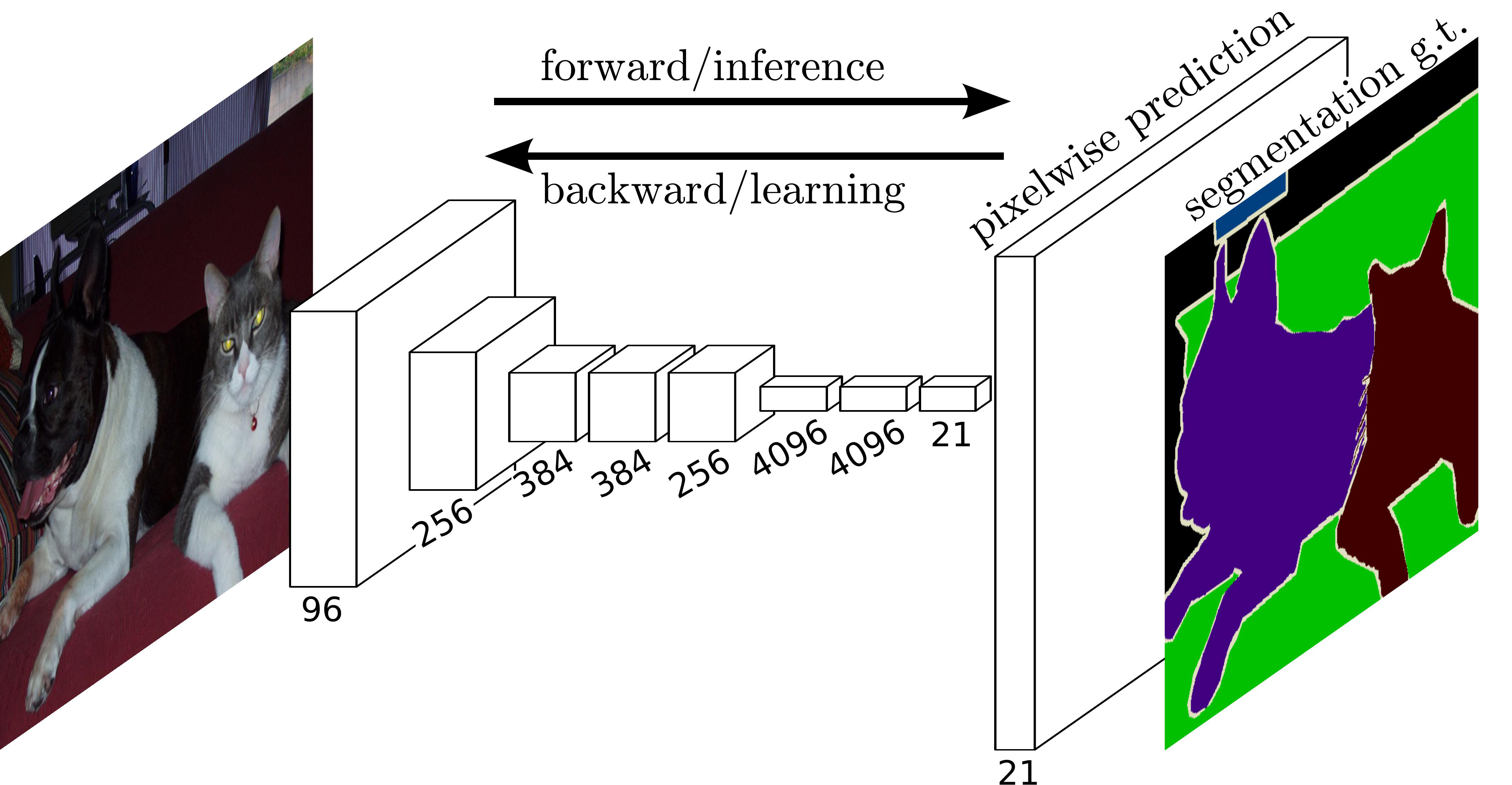}
    \caption{\textcolor{black}{Diagram showing the well-known Fully Convolutional Network (FCN) for segmentation tasks (Image from \cite{long2015fully}).}}
    \label{fig:seg_cnn}
\end{figure}

\textcolor{black}{\textbf{From classification to segmentation.}} Segmentation is a common task in both natural and medical image analysis which consists on assigning a class labels to each pixel in an image. A naive approach is to employ classification CNNs to classify each pixel in the image individually, by feeding it with patches extracted around each pixel. Nevertheless, an important shortcoming of this simple ‘sliding-window’ strategy is that input patches from neighboring pixels present a huge overlap and, as a result, the same convolutions are computed many times. Fully Convolutional Networks (FCNs) mitigate these limitations by treating the network as a single non-linear convolution, which can be trained end-to-end \cite{long2015fully} \textcolor{black}{(Fig. \ref{fig:seg_cnn} depicts the original FCN proposed in \cite{long2015fully})}. Unlike classification CNNs, FCNs are composed exclusively of convolutional blocks, which brings several benefits. First, they can be applied to images or arbitrary size, contrary to classification CNNs which require fixed input size. And second, since the predicted class score map is obtained in a single inference step, redundant convolution operations are avoided, resulting in more efficient architectures. More details about segmentation architectures are given in the next section. 


\section{A quick journey on medical image segmentation with deep learning}

The medical image computing community has greatly benefited from the pivotal developments in computer vision based on deep learning. It is undeniable that, nowadays, deep learning is dominating the literature in medical image processing, in challenging topics such as semantic segmentation. The seminal work in \cite{ronneberger2015u} extended the idea behind FCN \cite{long2015fully} and proposed UNet, whose design is still the backbone for many novel segmentation architectures in the medical field. This network basically consists on a contractive and expansive path. While the former is used to encode the input image into a compact latent representation, the latter upsamples these features to the original image size. Furthermore, the authors integrated skip-connections between the convolutional layers of the contracting and expanding paths, which help to fuse low and high-level features, as well as to facilitate the gradient flow through the network. 

Despite the fact that medical images are often in the form of 3D volumes, early use of CNNs in medical image segmentation resorted to slice-by-slice analysis \cite{prasoon2013deep,zhang2015deep}, particularly due to high complexity on initial 3D models. 
Nevertheless, an important limitation of this strategy is that the anatomic context in directions orthogonal to the 2D plane employed is completely discarded. To incorporate 3D contextual information while not incurring on costly operations, a common solution was to use 2D CNNs on images from the three orthogonal planes \cite{de2015deep,wang2017automatic}. Another alternative is to impose volumetric homogeneity on the individual 2D slices by constructing a 3D conditional random field (CRF) using scores from the CNN as unary potentials in a multi-label energy minimization problem \cite{shakeri2016sub}. With the advent of deep learning frameworks and the easy access to more powerful resources, researchers started to overcome the computational limitations and employ architectures with 3D convolutions \cite{kamnitsas2017efficient,dolz20183d,milletari2016v,fechter2017esophagus}. Nowadays, these 3D architectures can perform the segmentation of a whole MRI brain volume or a CT abdominal scan in few seconds.

The modular nature of CNN architectures makes them easily adaptable to multiple input sources of information. This is particularly important in medical imaging, where it is not uncommon to have multi-modal images from a given patient. Initial multi-modal CNN segmentation models adopted an early-fusion strategy, which integrated the multi-modality information from the original space of low- level features \cite{zhang2015deep,moeskops2016automatic}. Alternative works advocate that late fusion of high-level features is a better way to account for the complex relationships between multiple different modalities \cite{nie2016fully}. Nevertheless, fusing the multi-modal learned features either in an early or late step reduces the representation power of the network, since the modeling of the multiple modalities relies entirely on a single joint layer. To enhance the learning representation power, HyperDenseNet was proposed in \cite{dolz2018hyperdense}, where 
connectivity between modalities occur not only at a single level within the same path, but also between several levels across different paths. This grants the network with total freedom to learn more complex combinations between the modalities, increasing significantly the learning representation in comparison to early/late fusion. Following this intuition, several works to efficiently combine multiple image modalities have been proposed \cite{dolz2018ivd,multimodalKD,wang2019benchmark}.

As mentioned previously training a deep neural network results in optimizing a predefined objective function. As natural extension from classification models, segmentation networks have widely used the standard cross-entropy (CE) loss function, which is the de-facto solution in classification. Nevertheless, it is noteworthy to mention that in segmentation other metrics are typically privileged, such as the Dice score or the Hausdorff distance. As a result, objective functions have received significant attention recently, which can be categorized into: regional \cite{milletari2016v,wong20183d,sudre2017generalised,salehi2017tversky} and boundary losses \cite{kervadec2018boundary,karimi2019reducing}. The Dice overlap coefficient, or Dice score, was the first alternative to CE as loss function on segmentation networks \cite{milletari2016v}, typically outperforming CE in unbalanced problems \cite{wong20183d}. A generalized Dice loss was proposed in \cite{sudre2017generalised} by weighting according to the inverse of class-label frequency. To better balance the classes in terms of their relative class sizes, some researchers have investigated how to control the importance of false positives and false negatives \cite{salehi2017tversky}. A limitation of these regional losses is that they may encounter difficulties when dealing with very small structures, a common situation in imbalanced scenarios. In these cases, such regional losses have values that differ significantly across target classes --sometimes up to several orders of magnitude--, which may hamper the performance and training stability. On the other hand, boundary losses \cite{kervadec2018boundary,karimi2019reducing} can mitigate these issues since only the boundary between the regions is employed, instead of the whole region.



We want to emphasize that this is a quick overview of the status of medical image segmentation based on deep learning techniques. The objective of this section was to provide the reader with a global understanding and give a proper context of the field, and we do not intend that this represents a thorough literature review on this topic. For a more detailed survey on semantic segmentation methods in medical images we refer the reader to the great work in \cite{litjens2017survey}.

\section{Bladder cancer segmentation}

Knowing the precise location of the anatomical boundaries of inner-wall (IW), outer-wall (OW) and tumors is crucial to identify the tumor invasiveness. However, the automatic delineation of IW and OW in MRI images is a challenging task, due to important bladder shape variations, strong intensity inhomogeneity in urine caused by motion artifacts, weak boundaries and complex background intensity distribution \cite{duan2010coupled,qin2014adaptive,duan2012adaptive}. When further considering the presence of cancer, the problem becomes much harder as it introduces more variability across population. That might explain why literature on multi-region bladder segmentation remains scarce, with few techniques proposed to date (Table \ref{table:methods}). 

The trend in bladder segmentation is not different from what we have witnessed in other applications, where deep learning based methods prevail over classical approaches. We can observe that the number of papers based on traditional methods published in 14 years is lower than the deep learning based approaches in the period covering the last 4 years.

\subsection{Early attempts}

Early literature resorted to conventional computer vision approaches to address the problem of IW segmentation, such as Markov Random Fields \cite{li2004new,li2008segmentation} or region growing \cite{garnier2011bladder}. For example, in \cite{garnier2011bladder}, authors proposed to combine an active region growing strategy with a fast deformable model to overcome the leakage issue found in standard region growing methods. Particularly, their approach employing an inflation force, which acts like a region growing process, and an internal force that constrains the shape of the surface. Nevertheless, the direct application of these approaches to OW segmentation is impractical, \textcolor{black}{due} to the complex distribution of tissues surrounding the bladder. Furthermore, if the regions of interest contain large deformations, these methods present several important limitations. \textcolor{black}{For instance, large deformations may contain important noise or intensity variations, which will likely result in holes or oversegmentations. In addition, region growing is strongly sensitive to seeds placement, with different seeds potentially giving different segmentation results.}

\textcolor{black}{Mathematical morphology approaches have also been adopted for bladder segmentation. For example, Bueno et al.  \cite{bueno2001automatic} evaluated a method based on the watershed transformation to delineate pelvic structures in X-Ray CT images. Nevertheless, even though these methods can be quickly tuned and computed, they strongly depend on the quality of the image.}

Level-set based segmentation methods have also been proposed to segment both inner and outer bladder walls \cite{duan2010coupled,chi2011segmentation,han2013unified,qin2014adaptive,xiao20163d}. Duan et al. \cite{duan2010coupled} proposed a coupled level-set framework which integrated a modified version of the Chan-Vese model for IW and OW segmentation on T1-weighted MRI. In \cite{chi2011segmentation}, a model based on geodesic active contours was employed to segment the IW in T2-weighted MRI. The results from this step were then coupled with the constraint of maximum wall thickness in T1-weighted MRI images to segment the OW. Nevertheless, an important drawback of this method results from the difficulty to align corresponding slices between the two sequences. Xiao et al. \cite{xiao20163d} extended prior works based on level-sets by adding a second step, where fuzzy C-means was integrated in the pipeline to also segment tumour regions, in addition to IW and OW. Nevertheless, their results were not conclusive, since the results from this method were very inconsistent across datasets. Despite being a dominant approach in the past, however, level-sets approaches have several important shortcomings. First, as they are based on local optimization techniques, they are highly sensitive to initialization and image quality. Second, if the region of interest is embedded in another larger region, multiple manual initializations of the active contours are required, incurring in extra user interaction and larger processing times. Third, gaps in the target might result in evolving contours leaking into those gaps, leading to incomplete object contours. And fourth, processing times can be prohibitive, particularly in medical applications where the segmentation task is typically performed in volumetric fashion. Previous works have reported segmentation times usually exceeding 20 minutes for a single 3D volume.

\begin{table*}[h!]
\scriptsize
\centering
\begin{tabular}{l|ccc|ccc}
\toprule
   & \multicolumn{3}{c}{\textbf{Modality}}                                        & \multicolumn{3}{c}{\textbf{Target}}                                        \\
   \midrule
         & CT                   & \multicolumn{1}{c}{MR-T1} & \multicolumn{1}{c}{MR-T2} & IW                   & \multicolumn{1}{c}{OW} & \multicolumn{1}{c}{Tumour} \\
         \midrule
         \textbf{Pre deep learning era} \\
Li et al \cite{li2004new} (2004) & - & \checkmark& \checkmark& \checkmark & - & -\\
Li et al \cite{li2008segmentation} (2008) & - & \checkmark& \checkmark& \checkmark & - & -\\
Duan et al \cite{duan2010coupled} (2010) & -& \checkmark&- & \checkmark & \checkmark & -\\
Chi et al \cite{chi2011segmentation} (2011) & - & \checkmark& \checkmark& \checkmark & \checkmark & -\\
Garnier et al \cite{garnier2011bladder} (2011) & -& -&\checkmark & \checkmark & - & -\\
Ma et al \cite{ma2011novel} (2011) & - & -&  \checkmark & \checkmark & \checkmark & \checkmark\\
Chai et al \cite{chai2012automatic} (2012) & \checkmark & - & - & \checkmark & - & -\\

Duan et al \cite{duan2012adaptive} (2012) & -&\checkmark & -& - & - & \checkmark\\
Han et al \cite{han2013unified} (2013) & -& \checkmark& -& \checkmark & \checkmark & -\\
Qin et al \cite{qin2014adaptive} (2014) & -& -& \checkmark & \checkmark & \checkmark & -\\
Xiao et al \cite{xiao20163d} (2016) & -& -& \checkmark& \checkmark & \checkmark & \checkmark\\
Pinto et al \cite{pinto2017versatile} (2017) & \checkmark& -& -& \checkmark & - & -\\
Xu et al \cite{xu2017simultaneous} (2017) & -& -& \checkmark& \checkmark & \checkmark & -\\
Zhu et al \cite{zhu2018shape} (2018) & -&- & \checkmark& \checkmark & \checkmark & -\\

\midrule
         \textbf{Deep learning approaches} \\
Cha et al. \cite{cha2016bladderseg} (2016) & \checkmark & -  &  -  & \checkmark & -   & - \\ 
Cha et al. \cite{cha2016bladder} (2016) & \checkmark & -  &  -  & \checkmark &   - &  -\\ 

Gordon et al. \cite{gordon2017segmentation} (2017) & \checkmark & -  &  -  & \checkmark & \checkmark   & - \\

Dolz et al. \cite{dolz2018multiregion} (2018) & - & -  &  \checkmark  & \checkmark & \checkmark & \checkmark \\ 
Gsaxner et al. \cite{gsaxner2018pet} (2018) & \checkmark & -  &  -  & \checkmark & - & - \\ 
Leger et al. \cite{leger2018contour} (2018) & \checkmark & -  &  -  & \checkmark & - & - \\

Xu et al. \cite{xu2018automatic} (2018) & \checkmark & -  &  -  & \checkmark & - & - \\
Brion et al. \cite{brion2019using} (2019) & \checkmark & -  &  -  & \checkmark & - & - \\

Liu et al. \cite{liu2019bladder} (2019) & - & -  &  \checkmark  & \checkmark & \checkmark & \checkmark \\ 
Gordon et al. \cite{gordon2019deep} (2019) & \checkmark & -  &  -  & \checkmark & \checkmark & - \\
Hammouda et al. \cite{hammouda2019deep} (2019) & - & -  &  \checkmark  & \checkmark & \checkmark & \checkmark \\ 
Hammouda et al. \cite{hammouda2019cnn} (2019) & - & -  &  \checkmark  & \checkmark & \checkmark & - \\ 
Ma et al. \cite{ma20192d} (2019) & \checkmark & -  &  -  & \checkmark & - & - \\
Ma et al. \cite{ma2019u} (2019) & \checkmark & -  &  -  & \checkmark & - & - \\
Hammouda et al. \cite{hammouda20203d} (2019) & - & -  &  \checkmark  & \checkmark & \checkmark & \checkmark \\ 

\bottomrule

\end{tabular}
\caption{Literature on multi region bladder segmentation based on standard computer vision \textit{(top)} and deep learning \textit{(down)} approaches.}
\label{table:methods}
\end{table*}

As alternative to level-set approaches, Ma et al. \cite{ma2011novel} proposed to modify a a modified Geodesic active contour (GAC) and a shape-guided Chan-Vese model to segment bladder walls. More recently, a continuous max-flow framework with global convex optimization was proposed in \cite{xu2017simultaneous} to segment both IW and OW. 

Nevertheless, \textcolor{black}{these works present several important limitations. From an evaluation perspective, these works rarely report quantitative evaluation metrics, and they differ across works when they are provided. For example, while the Dice Similarity Coefficient (DSC) has been reported mostly reported \cite{qin2014adaptive,xu2017simultaneous,zhu2018shape}, other works resort to a more subjective evaluation, i.e., \textit{successful} vs \textit{unsuccessful} contours \cite{pinto2017versatile}, distance-based metrics \cite{chi2011segmentation,pinto2017versatile} or the impact on the conformity index \cite{chai2012automatic}. Furthermore, the limited size of the datasets, which typically range from 5 \cite{han2013unified,xu2017simultaneous} to 11 patients \cite{qin2014adaptive,xiao20163d,pinto2017versatile} impedes the evaluation of their generalization capabilities to larger cohorts. And second, from a methodological point of view} their high sensitivity to initialization makes the full automation of segmentation highly challenging for standard computer vision methods. In addition, another main limitation that hinders their usability in clinical context is that most methods focus only on bladder walls and are unable to segment simultaneously both bladder walls and tumors.  


\subsection{Deep learning era}

\textcolor{black}{Table \ref{table:closerlook} summarizes the current literature on bladder segmentation based on deep learning models. In this table, we highlight important information of each of these methods, such as architectural choices (e.g., network, 2D vs 3D), objective functions and employed image modality. In the following sections, these details are further expanded.}

\subsubsection{\textcolor{black}{Classification-based networks}}
Preliminary work adapted 2D classification networks to achieve the segmentation task \cite{cha2016bladderseg,cha2016bladder,gordon2017segmentation,gordon2019deep}. \textcolor{black}{Particularly, these studies employed the seminal work in \cite{krizhevsky2017imagenet}, AlexNet, as the backbone architecture for their approach. This network consists of 5 main layers, which include: 2 convolution layers with 64 kernels of size 5$\times$5, each followed by a pooling layer that reduces the output dimensionality, 2 locally-connected layers of size 64$\times$3$\times$3 and 32$\times$3$\times$3, and a final fully connected layer. Last, a softmax layer is added to transform the logits into probability values.} As discussed in Section \ref{sec:closerlook}, a main drawback of employing classification networks for segmentation is that pixel prediction is performed independently, hampering the spatial consistency in the prediction. To mitigate this issue, these works integrate a post-processing step, where level-sets \cite{osher1988fronts} are used to propagate the initial segmentation surface toward the bladder boundary. Nevertheless, in addition to the costly process of performing individual pixel predictions, the drawbacks of level-sets methods --i.e., sensitivity to initialization, image intensity and considerably slow process-- have already been detailed. These issues are further magnified in the case of tumor segmentation, due to its high shape and intensity variability. This may explain why prior works focused only on inner and outer wall, which makes them impractical to assess the muscle invasiveness in bladder cancer.

\begin{table*}[h!]
\scriptsize
\centering
\begin{tabular}{lccccc}
\toprule
 & \textbf{Architecture} & \textbf{Dimension} & \textbf{\begin{tabular}[c]{@{}c@{}}Number of patients\\
 (Train/Val/Test)\end{tabular}} & \multicolumn{1}{c}{\textbf{Loss}} & \textbf{Modality} \\
 \midrule
 Cha et al. \cite{cha2016bladderseg} &  Classification  &  2D   &  81/--/92   &   CE & CTU\\
 Cha et al. \cite{cha2016bladder} &  Classification  &  2D   &  62 ($n$-fold CV)   &   CE & CTU\\
 Gordon et al. \cite{gordon2017segmentation} &  Classification   &    2D     &    81/--/92  &     CE  & CTU  \\
 Dolz et al. \cite{dolz2018multiregion}  &   UNet-based     &  2D   &         40/5/15 (LOOCV)         &  CE & MR-T2 weighted\\
  Gsaxner et al. \cite{gsaxner2018pet}  &   FCN \cite{long2015fully}, ResNet \cite{he2016deep}   &  2D  &  29  & CE  & CT\\
 Leger et al. \cite{leger2018contour}  &   UNet   &  2D  &  179/80/80  & Dice & CT\\
 Xu et al. \cite{xu2018automatic} &   V-net \cite{milletari2016v}   &  3D   &   100/--/24     &  CE & CT \\
 Brion et al. \cite{brion2019using}  &  UNet  &  3D  &  96/8/8 ($n$-fold CV)  & Dice  & CT and CBCT\\
 Liu et al. \cite{liu2019bladder} &   UNet-based     &  2D   &   40/--/7 ($n$-fold CV)    & CE + Dice & MR-T2 weighted \\
Gordon et al. \cite{gordon2019deep} &  Classification & 2D  &   81/--/92    & CE & CTU \\
 Hammouda et al. \cite{hammouda2019deep} &  DeepMedic \cite{kamnitsas2017efficient}  &   2D &   20 (LOOCV)   & CE & MR-T2 weighted\\
  Hammouda et al. \cite{hammouda2019cnn} &  DeepMedic \cite{kamnitsas2017efficient}  &   3D &   10 (LOOCV)   & CE & MR-T2 weighted\\
  Hammouda et al. \cite{hammouda20203d} &  DeepMedic \cite{kamnitsas2017efficient}  &   3D &   17 (LOOCV)   & CE & MR-T2 weighted\\
  
Ma et al. \cite{ma20192d}  &  UNet  & 2D/3D &   74/7/92   &  CE & CTU\\
 Ma et al. \cite{ma2019u}  &  UNet  & 2D/3D &   74/7/92   & CE  & CTU \\

 \bottomrule
\end{tabular}
\caption{Closer look at deep learning methods on bladder segmentation.}
\label{table:closerlook}
\end{table*}


\subsubsection{\textcolor{black}{Fully-connected architectures}}
\begin{figure}
    \centering
    \includegraphics[width=1.0\linewidth]{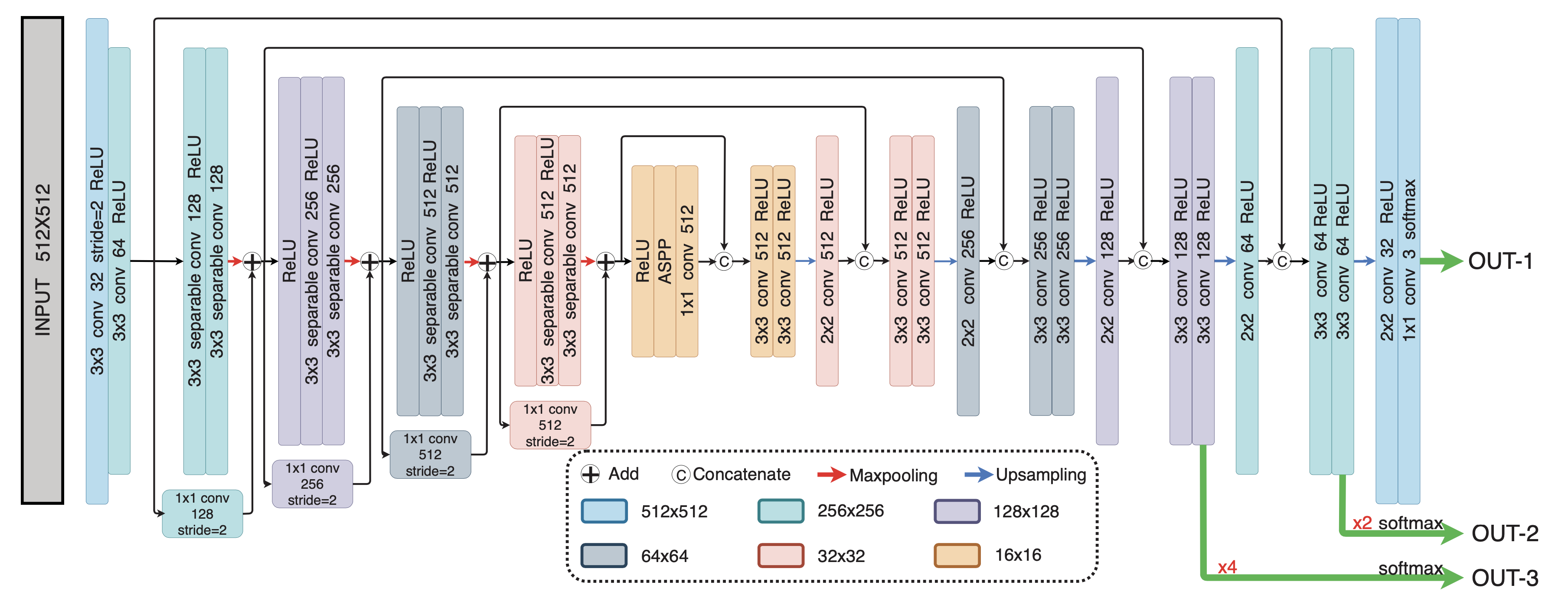}
    \caption{\textcolor{black}{The overall architecture proposed in \cite{liu2019bladder} (image from \cite{liu2019bladder}).}}.
    \label{fig:liu}
\end{figure}

Dolz et al. \cite{dolz2018multiregion} proposed the first work where a FCN was employed for multi-region bladder segmentation, including inner and outer wall, as well as tumor, in MR-T2 weighted images. To tackle with shape variability and strong intensity inhomogeneity, authors modified the well-known 2D UNet architecture by integrating progressive dilated convolutions. \textcolor{black}{Particularly, instead of gradually increasing the dilation factor through different convolutional layers, the authors increased this factor only within each context module. By doing this, the features learned at each block were able to capture multi-scale level information without incurring in additional costs in terms of learnable parameters.} This strategy allowed to span broader regions of input images, capturing more context while preserving the resolution of the analyzed regions. As tumor regions are unpredictable and they can be either very small or very large regions this method benefits to both cases. Following this work, similar approaches have been investigated under different conditions \cite{gsaxner2018pet,leger2018contour,liu2019bladder}. Gsaxner et al. \cite{gsaxner2018pet} compared to well-known architectures, i.e., FCN \cite{long2015fully} and ResNet \cite{he2016deep}, to segment the bladder in CT scans. \textcolor{black}{In their experiments, authors evaluated the impact of different data augmentations, such as rotation, scaling or zero-mean Gaussian-noise, on the final segmentation performance. From the reported results, authors observed that augmentations based on rotation and scaling alone brought larger improvements than augmentations based on these transformations and Gaussian-noise.} In \cite{leger2018contour}, authors employed the manual bladder segmentation on an adjacent slice as prior knowledge for the current slice. Thus, the network integrates prior knowledge in the form of an additional input channel, along with the current CT slice. \textcolor{black}{The backbone architecture employed was UNet with the default hyperparameters, with the only modification of the input size, in order to accommodate the target image and the segmentation of the previous slice.} However, this strategy is tailored to bladder segmentation, since the tumor shape significantly changes across the population. Thereby, incorporating shape prior will likely hamper the performance of the deep model. Similarly to \cite{dolz2018multiregion}, Liu et al \cite{liu2019bladder} proposed architectural modifications on UNet to accommodate large shape variations on the target in a multi-class segmentation scenario. In addition of dilated convolutions at multiple levels, they also incorporated multi-scale predictions. This sort of deep supervision brings several benefits. First, the predicted segmentation at different scales should be more consistent, leading to a better semantically representation of the targets. And second, it helps the convergence during training, as it facilitates the gradient flow and alleviates the problem of vanishing gradients. 
\textcolor{black}{The details of this architecture are depicted in Fig \ref{fig:liu}.}

Despite the satisfactory results obtained by previous methods, multi-region bladder segmentation in a slice-wise basis is not a trivial task. The large shape and size variability of bladder regions, the poor contrast between its wall and surrounding soft tissues and the presence, or not, or contrast material make of this a challenging task. Similarly to the literature on general medical image segmentation, recent works have investigated the impact of using 3D convolutional networks \cite{xu2018automatic,brion2019using,hammouda2019cnn,ma20192d,ma2019u}. For example, V-Net \cite{milletari2016v} was evaluated in \cite{xu2018automatic} in the context of bladder segmentation in CT. Particularly, authors employed the generated probability map to create bladder density maps. These maps were used as a second channel in the network input to further improve segmentation accuracy. Last, the authors incorporated a 3D fully connected CRF to refine the coarse voxel-wise probability maps generated by the model \textcolor{black}{and produce final fine-localized segmentation result, following prior works \cite{chen2017deeplab,kamnitsas2017efficient}. Concretely, the objective in a CRF model is to get the most probable label assignment $\mathbf{x}$ by minimizing an energy function of the form:}

\begin{equation}
    \label{eq:crf}
    E(\mathbf{x})= \sum_i \Psi_u (x_i) + \sum_{i,j} \Psi_p (x_i,x_j)
\end{equation}

\textcolor{black}{This function is composed of unary ($\Psi_u$) and pairwise ($\Psi_p$) potentials. The unary potential can be computed from the CNN output label probabilities at pixel $i$ as $\Psi_u (x_i) = - \log(s_i)$. On the other hand, following the work in \cite{krahenbuhl2011efficient}, the pairwise potential can be defined as:} %

\begin{equation}
    \label{eq:crf-pair}
    \Psi_p (x_i,x_j) = \mu (x_i,x_j) \left[w_1 \exp \large(- \frac{\|p_i-p_j\|^2}{2 \sigma_{\alpha}^2}- \frac{\|I_i-I_j\|^2}{2 \sigma_{\beta}^2}) + w_2 \exp \large(- \frac{\|p_i-p_j\|^2}{2 \sigma_{\lambda}^2})\right]
\end{equation}

\textcolor{black}{where $\mu (x_i,x_j)=1$ if $x_i \neq x_j$ and 0 otherwise. The expression in \ref{eq:crf-pair} uses two Gaussian kernels in different feature spaces. The first kernel depends on pixel positions, defined as $p$, and on color intensities, referred to as $I$, whereas the second kernel only depends on pixel positions. The hyperparameters $w_1$ and $w_2$ balance the importance of the two terms, while $\sigma_{\alpha}$, $\sigma_{\beta}$ and $\sigma_{\gamma}$ control the scale of Gaussian kernels. The intuition behind these kernels is that while the first one forces pixels having similar position and colors to have the same labels, the second kernel only considers spatial information to enforce smoothness.}

More recently Brion et al. \cite{brion2019using} used a 3D-UNet \cite{cciccek20163d} coupled with a Dice score loss \cite{milletari2016v} to segment the bladder on cone beam CT (CBCT) images. \textcolor{black}{Particularly, the Dice score coefficient (DSC) between two binary volumes A and B can be defined as:}

\begin{equation}
    \label{eq:diceloss}
\mathrm{DSC} \ = \  \frac{2 \sum_{i}^N a_i b_i} {\sum_{i}^N a_i^2 + \sum_{i}^N b_i^2}
\end{equation}

\textcolor{black}{where $N$ is the total number of voxels, and $a \in A$ and $b \in B$ are the sets of voxels in the predicted segmentation and its corresponding ground truth. Then, we can compute its derivative with respect to the $j$-\textit{th} voxel of the predicted volume $A$, which results in the following gradient: }

\begin{equation}
    \label{eq:dicelossgrad}
\frac{\partial DSC}{\partial a_j}  =  2 \left[  \frac{b_j (\sum_{i}^N a_i^2 + \sum_{i}^N b_i^2) - 2 a_j (\sum_{i}^N a_i b_i)} {(\sum_{i}^N a_i^2 + \sum_{i}^N b_i^2)^2} \right]
\end{equation}

\begin{figure}
    \centering
    \includegraphics[width=1.0\linewidth]{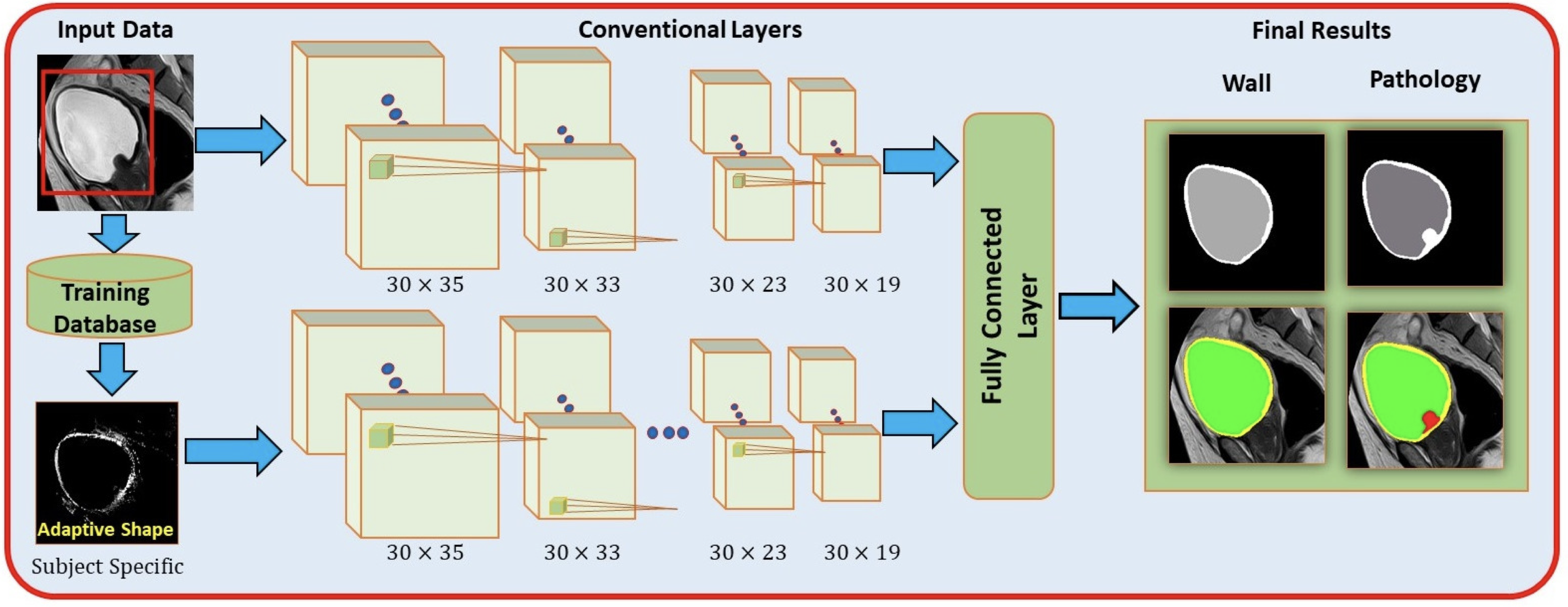}
    \caption{\textcolor{black}{Schematic illustration of the CNN approach proposed in \cite{hammouda2019deep} for bladder-wall segmentation (Image from \cite{hammouda2019deep}).}}
    \label{fig:priorshape}
\end{figure}

Another well-known architecture in the medical community, i.e., DeepMedic \cite{kamnitsas2017efficient}, was evaluated in \cite{hammouda2019deep,hammouda2019cnn} to segment the inner and outer wall in MR-T2 weighted images. \textcolor{black}{In addition to the input MRI, authors in \cite{hammouda2019deep} incorporate shape prior information derived from the whole training set (Fig. \ref{fig:priorshape}). To achieve this, MRI and ground-truth images were registered by applying an affine transformation, which is followed by a B-splines based transformation \cite{glocker2011deformable}. The resulting transformation parameters are employed to generate the individual shape prior probability that is coupled with the MRI to form the network input. Nevertheless, this presents two limitations. First, both training and testing images need a pre-processing step to align the images. And second,} since authors use additional shape priors \cite{hammouda2019deep} during training, this approach is also limited to the bladder. More recently, this idea was further extended in \cite{hammouda20203d} to also account for tumor regions. Particularly, authors proposed to employ the network in \cite{kamnitsas2017efficient} with two branches, where the first one captures bladder wall and tumor as a single target, whereas the second network segments the boundaries simulating a bladder without pathology. Nevertheless, this approach requires building an active shape prior which depends on pre-registration steps. Furthermore, it might be highly sensitive to several hyperparameters, such as the number of shapes to integrate in the shape prior model or the hyperparameters of the fully dense CRF employed as post-processing.

\subsubsection{\textcolor{black}{Evaluation on the literature}}

\paragraph{\textcolor{black}{\textbf{Evaluation metrics}}}

\textcolor{black}{Ideally, to foster the progress in this task and conduct appropriate experiments with accurate comparisons, the existing models should be evaluated under similar metrics. Nevertheless, we have found that this is an underlying issue in the bladder segmentation literature, since there is no a common evaluation metric across the different works. Therefore, direct comparison between the existent work is a hard task. We present below the metrics employed in the analyzed works.}

\textcolor{black}{The Dice coefficient, or Dice similarity score (DSC), is the most popular metric across the literature \cite{dolz2018multiregion,gsaxner2018pet,leger2018contour,xu2018automatic,brion2019using,hammouda2019cnn,hammouda2019deep,hammouda20203d}, which is commonly used in many medical image segmentation problems. This metric can be defined as twice the overlap area of predicted and ground-truth maps, divided by the total number of pixels in both images. The formula to compute the DSC for two volumes A and B is given in Eq. \ref{eq:diceloss}.}


\textcolor{black}{The Jaccard Index (JI), which is similar to DSC, has also been widely employed to evaluate bladder segmentation methods \cite{cha2016bladder,leger2018contour,brion2019using,ma20192d,ma2019u}. The JI can be computed as follows:}

\begin{equation}
    \label{eq:ji}
\mathrm{JI(A, B)} \ = \  \frac{ A \cap B} { A \cup B}
\end{equation}

\textcolor{black}{In addition to these well-known region-based metrics, other less-employed criteria include the volume intersection (VI) between ground truth and predicted volumes \cite{cha2016bladderseg,gordon2017segmentation,ma20192d}, volume error (VE) \cite{cha2016bladderseg,gordon2019deep,ma20192d}, and relative volume difference (RVD) \cite{xu2018automatic}. }

\textcolor{black}{Nevertheless, region or volume based metrics typically lack sensitivity to segmentation outline, which might result in segmentations with high degree of spatial overlap presenting clinically-
relevant differences between their contours. Thus, distance-based metrics are sometimes included in the evaluation of proposed methods. For example, the average symmetric surface distance (ASSD) has been considered in \cite{cha2016bladderseg,dolz2018multiregion,brion2019using,gordon2019deep}. The ASSD between contours $A$ and $B$ is defined as follows:}

\begin{equation}
\mathrm{ASSD(A, B)} \ = \ \dfrac{1}{|A| + |B|} \, \left(\sum_{a \in A} \min_{b \in B} \, d(a,b) + \sum_{b \in B} \min_{a \in A} \, d(b,a) \right),
\end{equation}

\textcolor{black}{where $d(a,b)$ is the distance between point $a$ and $b$.}

\textcolor{black}{The Hausdorff distance (HD), which measures the maximum distances between the closest points of two contours, is also a popular distance-based evaluation metric \cite{gsaxner2018pet,hammouda2019cnn,hammouda2019deep,hammouda20203d,ma2019u}. It can be defined as:}

\begin{equation}
\mathrm{HD(A, B)} \ = \ \max \Big\{ \max_{a \in A}\{\min_{b \in B}\{d(\mathbf{a},\mathbf{b})\}\}, \max_{b \in B}\{\min_{a \in A}\{d(\mathbf{a},\mathbf{b})\}\}\Big\},
\end{equation}

\paragraph{\textcolor{black}{\textbf{Evaluation strategies}}}
\textcolor{black}{Another factor that strongly impacts the performance of a deep learning model is the size of the training and testing datasets, as well as the evaluation strategy. First, regarding the number of training and testing images, we have identified three main groups: \textit{i}) less than 50 patients \cite{gsaxner2018pet,brion,liu2019bladder,hammouda2019cnn,hammouda2019deep,hammouda20203d}, \textit{ii} between 50 and 100 patients \cite{gordon2017segmentation,dolz2018multiregion} and \textit{iii}) more than 100 patients \cite{cha2016bladder,leger2018contour,xu2018automatic,gordon2019deep,ma20192d,ma2019u}. This results in small test groups to evaluated the proposed methods, some including less than 10 testing images. While the limitation on the number of testing subjects can be alleviated by cross-validation, most methods, however, simply use a 1-fold validation strategy. An important issue with these practices is that the generability of the approaches is not fully demonstrated, as the testing set is relatively small. This is further aggravated in several approaches \cite{cha2016bladder,gordon2017segmentation,gsaxner2018pet,liu2019bladder}, which do not employ an independent validation set to fix the hyper-parameters or as an early stopping criteria.}

\begin{table}[h!]
\scriptsize
\centering
\begin{tabular}{lcccc}
\toprule
        & \multirow{2}{*}{\textbf{\begin{tabular}[c]{@{}c@{}}Nr of patients\\
 (Train/Val/Test)\end{tabular}}} & \multirow{2}{*}{Evaluation} & \multicolumn{2}{c}{Results (DSC)} \\
        &                           &                             & OW & Tumour \\
        \midrule

    Dolz et al. \cite{dolz2018multiregion}  &    40/5/15 & $1$-fold &  0.839 & 0.686   \\
     Dolz et al. \cite{dolz2018multiregion}*  &    40/--/7 & $1$-fold &  0.856 & 0.924   \\
 Liu et al. \cite{liu2019bladder} &    40/--/7 & $n$-fold CV  & 0.888&   0.954   \\
 Gordon et al. \cite{gordon2019deep} &  81/--/92  & $1$-fold  &  0.872  & -- \\
  Hammouda et al. \cite{hammouda2019deep} & 20 & LOOCV & 0.978 & 0.971  \\
     Hammouda et al. \cite{hammouda20203d} &   17 & LOOCV  &  0.974 &  0.957  \\
 \bottomrule
  \multicolumn{4}{l}{\scriptsize{* Results reported in \cite{liu2019bladder}.}}
\end{tabular}
\caption{\textcolor{black}{Summary of the performance of the proposed methods in the literature for outer wall (OW) and bladder tumor segmentation by employing the DSC metric.}}
\label{table:metrics2}
\end{table}

\paragraph{\textcolor{black}{\textbf{Quantitative results}}}

\textcolor{black}{Tables \ref{table:metrics2} and \ref{table:metrics1} report the numerical evaluation of the existing literature. Since the evaluation protocols, metrics and datasets employed to evaluate the current methods significantly differ across works, it is hard to provide a fair and systematic comparison. In terms of DSC, the approaches that have modified well-known architectures to accommodate for the challenges of bladder segmentation (i.e., \cite{dolz2018multiregion}), as well as those adopting 3D architectures \cite{hammouda20203d} have achieved the highest performance on the IW segmentation task. Furthermore, for the same target, Hammouda et al. \cite{hammouda2019deep} demonstrated that by incorporating anatomical shape prior during training, the value of distance-based metrics, such as MHD, can be considerably reduced. On the other hand, the scarce literature that has focused on OW and tumour segmentation, have resorted mainly to DSC as evaluation metric. Thus, we can observe that recent works \cite{liu2019bladder,hammouda2019cnn,hammouda20203d} achieve DSC values above 0.90 for both OW and tumour. Nevertheless, it is noteworthy to mention that differences across these works are likely affected by the characteristics of the dataset chosen, i.e., quality and quantity, as well as the evaluation protocol followed in the experiments.}

\begin{table*}[]
\scriptsize
\centering
\begin{tabular}{lccccccccc}
\toprule
        & \multirow{2}{*}{\textbf{\begin{tabular}[c]{@{}c@{}}Nr of patients\\
 (Train/Val/Test)\end{tabular}}} & \multirow{2}{*}{\textbf{Evaluation}} & \multicolumn{6}{c}{\textbf{Results (IW)}} \\ 
        &    &     & \begin{tabular}[c]{@{}c@{}}DSC  \\($\%$)\end{tabular}  & \begin{tabular}[c]{@{}c@{}}JI \\($\%$)\end{tabular}    & \begin{tabular}[c]{@{}c@{}}VI\\($\%$)\end{tabular} & \begin{tabular}[c]{@{}c@{}}VE\\($\%$)\end{tabular} & \begin{tabular}[c]{@{}c@{}}RVD\\($\%$)\end{tabular} & \begin{tabular}[c]{@{}c@{}}HD\\(mm)\end{tabular} & \begin{tabular}[c]{@{}c@{}}ASSD\\(mm)\end{tabular}      \\
        \midrule
 Cha et al. \cite{cha2016bladderseg}  &  81/--/92 & 1-fold & --& 0.726 &  0.819 & 0.102 & -- & -- & 3.6 \\
   Gordon et al. \cite{gordon2017segmentation} &  79/--/15  & $1$-fold &-- & -- & 0.876  &-- &-- &-- &-- \\
    Dolz et al. \cite{dolz2018multiregion}  &    40/5/15 & $1$-fold & 0.984 & -- & -- & -- & -- & -- & -- \\
   Gsaxner et al. \cite{gsaxner2018pet} (FCN)  &   21/--/8   &  $1$-fold  &   0.804 & -- &  --& -- & -- &  6.1 & --\\
      Gsaxner et al. \cite{gsaxner2018pet} (ResNet)  &   21/--/8   &  $1$-fold  &   0.769 & -- & -- & -- & -- &  6.3 & -- \\
  Leger et al. \cite{leger2018contour}  &   179/80/80  & $1$-fold & 0.919 & 0.861 & -- &  -- & -- &-- & -- \\
  Xu et al. \cite{xu2018automatic} &    100/--/24 & $1$-fold & 0.922 & -- & -- &  0.144 & -3.4 &--& 2.02 \\
        
 Brion et al. \cite{brion2019using}  & 32/8/8 & $6$-fold CV & 0.801 & 0.685 & -- & -- & -- & -- & --\\
 Liu et al. \cite{liu2019bladder} &    40/--/7 & $n$-fold CV  & 0.888& --&--&--& -- & --  & --\\
 Gordon et al. \cite{gordon2019deep} &  81/--/92  & $1$-fold  & -- & -- & 0.872 &-- &-5.3 & -- &  3.2 \\
  Hammouda et al. \cite{hammouda2019deep} & 20 & LOOCV & 0.990 & -- 
 & -- & -- &  -- &0.17 & --  \\
   Hammouda et al. \cite{hammouda2019cnn} &   10 &LOOCV  & 0.992 & -- & -- & -- &  -- & 0.40 &  --\\
     Hammouda et al. \cite{hammouda20203d} &   17 & LOOCV  & 0.980 & -- & -- & -- & -- & 0.13 &--  \\
 Ma et al. \cite{ma20192d} (2D) &   74/7/92 & $1$-fold &  --  &0.845&  0.929 & -- & -3.1 &-- &2.8   \\
 Ma et al. \cite{ma20192d} (3D) &   74/7/92 & $1$-fold &  --  &0.819&  0.913& -- & -4.1  & -- & 3.2 \\
  Ma et al. \cite{ma2019u} (2D) &  74/7/92 & $1$-fold &  --  &0.840&  0.936 &  -- & -5.7   & 10.3 & 2.9 \\
  Ma et al. \cite{ma2019u} (3D) &  74/7/92 & $1$-fold & -- & 0.811 & 0.901 &  -- & -3.1 & 11.5 & 3.3 \\
 
 \bottomrule
\end{tabular}
\caption{\textcolor{black}{Summary of the performance of the proposed methods in the literature for inner wall (IW) segmentation by employing several evaluation metrics.}}
\label{table:metrics1}
\end{table*}

\subsubsection{\textcolor{black}{Reproducibility}}

\textcolor{black}{Reproducibility is an essential requirement for studies based on deep learning techniques. In order for a study to be reproduced, an independent researcher requires information about the experimental setting, which includes datasets, network hyperparameters, and software or libraries employed, among others. Furthermore, it is also desirable that authors publicly share their code to the community. Nevertheless, we found that most recent literature on bladder segmentation based on deep learning does not provide sufficient information to enable reproducibility of the experiments. For example, the works in \cite{gordon2019deep,hammouda2019deep,hammouda2019cnn,ma20192d,hammouda20203d} lack important details to reproduce the results, such as the optimizer employed to update the network parameters, initial learning rate, number of training epochs or batch size, among others. Indeed, among all the papers analyzed in this review, only 5 mentioned the library used for their implementation, and the code is share in only 3 papers (details are given in Table \ref{table:code}). This limits the advancement of such results by independent researchers, as well as impedes the comparison of novel approaches with existing methodologies. }

\begin{table*}[h!]
\scriptsize
\centering
\begin{tabular}{lcc}
\toprule
 &  \textbf{Library} & \textbf{Code link}  \\
 \midrule
 Cha et al. \cite{cha2016bladderseg} &  -- & --  \\
 Cha et al. \cite{cha2016bladder} &  -- & --  \\
 Gordon et al. \cite{gordon2017segmentation} & -- & -- \\
 Dolz et al. \cite{dolz2018multiregion}  &  Pytorch 0.4.0   & \href{https://github.com/josedolz/Progressive_Dilated_UNet}{https://github.com/josedolz/Progressive$\_$Dilated$\_$UNet}      \\
  Gsaxner et al. \cite{gsaxner2018pet}  & TensorFlow 1.3 & \href{https://github.com/cgsaxner/UB_Segmentation}{https://github.com/cgsaxner/UB$\_$Segmentation} \\
 Leger et al. \cite{leger2018contour}  & Keras (TensorFlow) & -- \\
 Xu et al. \cite{xu2018automatic} & Caffe & \href{https://github.com/superxuang/caffe_3d_crf_rnn}{https://github.com/superxuang/caffe$\_$3d$\_$crf$\_$rnn}\\
 Brion et al. \cite{brion2019using}  & -- & --\\
 Liu et al. \cite{liu2019bladder} &  -- & --\\
Gordon et al. \cite{gordon2019deep} & -- & --\\
 Hammouda et al. \cite{hammouda2019deep} & -- & --\\
  Hammouda et al. \cite{hammouda2019cnn} &-- & --\\
Ma et al. \cite{ma20192d}  & -- & -- \\
 Ma et al. \cite{ma2019u}  & Keras (TensorFlow) & -- \\
   Hammouda et al. \cite{hammouda20203d} & -- & --\\

 \bottomrule
\end{tabular}
\caption{\textcolor{black}{Details of the libraries employed for implementation, and the public available codes, when available.}}
\label{table:code}
\end{table*}

\subsubsection{\textcolor{black}{Comparison to previous works}}

\textcolor{black}{As mentioned in the previous section, the lack of information to reproduce prior work, together with no public datasets to benchmark proposed methods, make really hard the comparison of novel approaches to existing literature. While some authors have compared to vanilla versions of the proposed architectures to demonstrate the effectiveness of their contributions \cite{dolz2018multiregion,leger2018contour,xu2018automatic,gordon2019deep,hammouda20203d}, some other employ weaker models designed for other purposes \cite{hammouda2019deep,hammouda2019cnn} or image registration approaches \cite{brion2019using}, compare to their own works \cite{ma2019u} or simply do not compare to any other method \cite{cha2016bladder,cha2016bladderseg,ma20192d}. In contrast, only the work in \cite{liu2019bladder} compares to both standard baselines and prior work, e.g., \cite{dolz2018multiregion}, in the experiments. The lack of comparisons to prior solutions might cast doubts about the real contributions of novel approaches.}

\section{Limitations \textcolor{black}{and potential directions}}

Even though the literature on bladder segmentation based on deep learning is slowly evolving, we have identified several important limitations. Our intention is that researchers can benefit from our findings by addressing the shortcomings of current approaches. 

In terms of methodology, nearly all the works simply employ standard architectures, which were designed for different purposes, i.e., UNet (\cite{leger2018contour,brion2019using,ma2019u,ma20192d}), FCN (\cite{gsaxner2018pet}), V-Net (\cite{xu2018automatic}) or DeepMedic (\cite{hammouda2019deep,hammouda2019cnn,hammouda20203d}). While we consider that evaluating the value of current research is essential, these models might not be optimized for certain targets, such as tumors. This was demonstrated by Dolz and Liu \cite{dolz2018multiregion,liu2019bladder}, who adapted standard neural networks to the specificities of the problem. Thus, we believe that specific architectural designs to accommodate, for example large shape or appearance variations, can bring novel higher performing models. In the same vein, categorical cross-entropy has been widely considered in prior work as objective function, with just a handful of papers employing the dice loss as the objective to optimize. Taking into account that multi-region bladder segmentation is a highly imbalanced setting, these losses may not be well suited for this problem. \textcolor{black}{For example, although Dice loss might work well on unbalanced scenarios, its gradient (Eq. \ref{eq:dicelossgrad}) has squared terms in the denominator. This makes that when those values are small, it could result in large gradients, which leads to training instability.} Current literature points out to regional losses, i.e., dice loss, underperforming boundary losses \cite{kervadec2018boundary,karimi2019reducing} when target class distributions are imbalanced. This is the case of bladder tumor segmentation, where just a few voxels might represent the tumour. Following these recent findings, adding these boundary loss terms on the objective function can bring a boost to the segmentation performance, particularly on the tumor regions.

Medical image segmentation in a multi-modal scenario has received a substantial research attention in important applications such as neuroimaging \cite{kamnitsas2017efficient,dolz2018hyperdense,wang2019benchmark}. In this setting, different image modalities are typically combined to overcome the limitations of individual imaging techniques. As demonstrated on recent works, processing multiple images simultaneously has a positive impact on the performance of deep models. This is explained by the combination of the statistical properties between image modalities, which can enhance the representation power of deep learning models. While simply merging multiple image modalities has demonstrated to significantly improve the performance of segmentation models based on single modalities, recent fusion strategies have achieved encouraging results on this scenario. Hence, we encourage the exploration of deep models that efficiently fuse information contained in multiple image modalities. Particularly, in the context of bladder tumor assessment, this results in leveraging multiparametric magnetic resonance (mp-MR) images, which is replacing CT for bladder cancer staging in clinical practices.

Another important limitation that can explain the reduced literature on this topic is the scarcity of available datasets. We understand that obtaining manual label annotations to train a deep model is a cumbersome process. Furthermore, ethical and privacy reasons impede the public availability of clinical datasets. Nevertheless, we have witnessed the fast growing literature on semantic segmentation applications that publicly shared labeled datasets, typically through international challenges in the medical image computing conferences. For example, brain tumor and tissue segmentation on multiple MRI, for which there exist an important body of literature, provide extensive labeled datasets through their challenges (BRATS\footnote{http://braintumorsegmentation.org}, iSEG\footnote{http://iseg2017.web.unc.edu} or MRBrainS \footnote{https://mrbrains18.isi.uu.nl}, to name a few). This can, in addition, provide an excellent environment to compare methods developed by researchers world-wide under the same conditions. This is particularly important to assess the real value of proposed methods. Up to date, results are rarely comparable, since each setting (e.g., number of patients for training/testing, hyperparameters, etc) is completely different. Furthermore, in most cases, neither the code or important implementation details are given \cite{hammouda2019deep,hammouda2019cnn,ma20192d,ma2019u,hammouda20203d}, which makes the reproducibilty of the results a challenging task.



Despite the impressive performance of deep networks in medical image segmentation, \textcolor{black}{there exist two common problems that impede their scalability. First, deep learning models tend to under-perform when they are trained on a dataset with an underlying distribution different from the target images. This may happen, for instance, when images are acquired across different scanner vendors or with different acquisition parameters. One solution to circumvent this problem is unsupervised domain adaptation, where only images in the source dataset are labeled. To leverage unlabeled target domain data, literature has typically employed adversarial training \cite{dou2019pnp}. In these approaches, representations of both source and target domain are forced to match at either the input or the feature space. Other solutions include direct matching in the network-output space \cite{pichler2020direct} or are driven by weakly annotations, for example in the form of size priors \cite{bateson2019constrained,bateson2020source}. These strategies have shown promising results in other medical problems and we believe their application on bladder cancer segmentation could be beneficial to assess the domain shift in this scenario.} And second, these models require for huge amount of labeled training data. From a human perspective, obtaining such manual labels is an extremely costly process, which is in addition prone to variability. An appealing alternative to alleviate this issue is to train segmentation networks under reduced supervision. For example, weakly supervised segmentation models can employ image-level tags \cite{adiga2020manifold,belharbi2020deep}, scribbles \cite{lin2016scribblesup,tang2018regularized}, bounding boxes \cite{kervadec2020bounding,rajchl2016deepcut} or global constraints \cite{kervadec2019curriculum,kervadec2019constrained,peng2020discretely}
as supervisory signals. In the context of medical imaging, Kervadec et al. \cite{kervadec2019constrained} proposed to employ global anatomical priors, in the form of a high-order loss that constrains the size of the target segmentation. With only a fraction of annotated pixels, the proposed size loss achieved a performance close to full supervision (i.e., all pixels labeled) on cardiac and prostate imaging.

\textcolor{black}{Last, the final goal of multiregion segmentation in bladder cancer is to assist on the staging of bladder cancer, i.e., MIBC vs NMIBC. This is of vital importance, as urothelial cancers become more aggressive when they gradually grow into and through the bladder wall. Nevertheless, none of existing segmentation works have addressed this important issue. We encourage further research to assess the performance of segmentation methods not only from a pixel-wise classification perspective, but from an image-level point of view. This would ideally highlight the real value of these methods in clinical routine for assessing bladder cancer patients.}

\section{Conclusion}

Multi region bladder segmentation is of pivotal importance in the prognosis of bladder cancer, as it can serve to stage the primary tumor. Nevertheless, even though deep learning models have achieved outstanding performance in many visual recognition tasks, including medical imaging segmentation, the literature on multi region bladder segmentation remains still scarce. In this survey, we have \textcolor{black}{delved into the details of current literature on this topic} and identified several limitations, which include: lack of architectures designed to address the specific challenges of this task, rich information on multimodal data is disregarded and current advances in the deep learning literature are not integrated. With this work, we expect to have shed light about the current status of multi-region bladder segmentation approaches based on deep learning and have given valuable insights on better practices to deploy higher performing models.


\bibliography{mybibfile_red}

\end{document}